\documentclass[runningheads]{llncs}

\usepackage{array}
\usepackage{amsmath}
\usepackage{amsbsy}
\usepackage{amssymb}
\usepackage{algorithm}
\usepackage{algpseudocode}
\usepackage{float}
\usepackage{listings}
\usepackage{url}
\usepackage{hyperref}
\usepackage{graphicx}
\usepackage{todonotes}
\usepackage{enumitem}

\usepackage{tikz-cd}

\usepackage{multicol,caption}

\usepackage[authoryear,round,sectionbib]{natbib}

\floatstyle{ruled}
\newfloat{algorithm}{tbp}{loa}
\providecommand{\algorithmname}{Algorithm}
\floatname{algorithm}{\protect\algorithmname}

\newcommand{\dataset}{\mathbf{X}}
\newcommand{\labels}{\mathbf{y}}
\newcommand{\nsamples}{N}
\newcommand{\nfeatures}{F}
\newcommand{\coreset}{\mathbf{T}}
\newcommand{\maxcoresamples}{M}

\newcommand{\traindata}[1]{\mathbf{X}^{tr}_{#1}}
\newcommand{\testdata}[1]{\mathbf{X}^{te}_{#1}}

\newcommand{\x}{\mathbf{x}_i}
\newcommand{\xI}[1]{\mathbf{x}_{#1}}
\newcommand{\y}{{y}_i}

\newcommand{\weights}{\mathbf{w}}
\newcommand{\weightI}[1]{{w}_{#1}}

\newcommand{\param}{\theta}
\newcommand{\paramI}[1]{\theta_{#1}}

\newcommand{\likelihood}[1]{\mathcal{L} \left( #1 \right)}
\newcommand{\prior}[1]{\pi_0 \left( #1 \right)}
\newcommand{\posterior}[1]{\pi \left( #1 \right)}

\newcommand{\lldata}{\log \mathcal{L} \left( \dataset; \param \right)}

\newcommand{\llcoreset}{\log \mathcal{L} \left( \coreset; \weights, \param  \right)}

\newcommand{\logistic}[1]{\sigma\left(#1\right)}
\newcommand{\sign}[1]{\textnormal{sign}\left(#1\right)}
\newcommand{\one}[1]{1\left[#1\right]}

\newcommand{\gaussianpdf}[2]{\mathtt{N}\left(#1,#2\right)}
\newcommand{\bernoullipdf}[1]{\mathtt{Bern}\left(#1\right)}

\begin{document}

	\title{Analyzing and Storing Network Intrusion Detection Data using Bayesian Coresets: \\ A Preliminary Study in Offline and Streaming Settings}
	\titlerunning{Analyzing Network Intrusion Detection Data using Bayesian Coresets}

	\author{Fabio Massimo Zennaro\orcidID{0000-0003-0195-8301}}
	\authorrunning{F.M. Zennaro}
	%
	\institute{Department of Informatics, University of Oslo, \\
		PO Box 1080 Blindern, 0316 Oslo, Norway \\
		\email{fabiomz@ifi.uio.no}\\}
	\maketitle              
	\begin{abstract}
		In this paper we offer a preliminary study of the application of Bayesian coresets to network security data. Network intrusion detection is a field that could take advantage of Bayesian machine learning in modelling uncertainty and managing streaming data; however, the large size of the data sets often hinders the use of Bayesian learning methods based on MCMC. Limiting the amount of useful data is a central problem in a field like network traffic analysis, where large amount of redundant data can be generated very quickly via packet collection. Reducing the number of samples would not only make learning more feasible, but would also contribute to reduce the need for memory and storage. We explore here the use of Bayesian coresets, a technique that reduces the amount of data samples while guaranteeing the learning of an accurate posterior distribution using Bayesian learning.  We analyze how Bayesian coresets affect the accuracy of learned models, and how time-space requirements are traded-off, both in a static scenario and in a streaming scenario.
		
	\end{abstract}

\section{Introduction \label{sec:Introduction}}

Securing modern networks is a non-trivial challenge that requires high throughput (to evaluate in real-time the behavior of the network) and expert knowledge (to decide whether suspicious or malicious activity is taking place on the network). Network intrusion detection, in particular, is concerned with the early detection of attempts of breaking into and/or comprising a network. Packet collection techniques allows the continuous monitoring of networks and the collection of large amounts of traffic data. 

Modern network intrusion detection data sets quickly exceeds the processing capacity of human reviewers, and thus automatic processing techniques for filtering and analyzing the data become necessary. 
A simple solution consists in eliciting knowledge from experts and encoding it into logical rules. Such rule-matching or signature-matching algorithms can process the data quickly, but their effectiveness is limited by the sort and the range of knowledge provided by the experts; in particular, these algorithms lack any sort of generalization and are unable to deal with novel data that do not perfectly match the encoded rules \citep{gardiner2016security}.
Machine learning has been put forward as a potential solution to this shortcoming. By learning directly from the data, and not from the particular knowledge provided by experts, machine learning algorithms aim at learning patterns that may hold not only for historical collected data, but also for future unforeseen data. 

Many current machine learning techniques rely on large amount of data to learn useful patterns. The success of \emph{deep learning}, in particular, has been explained by, among other factors, the availability of big data sets \citep{lecun2015deep}. However, large data sets have also significant drawbacks. Large collections of data are problematic to archive and to store on drives; they are challenging to manipulate and load, requiring either a high amount of memory or frequent swapping operations; they are often redundant, to the point that this redundancy contributes little or nothing to the learning process. 

Beyond deep learning, other machine learning algorithms may be severely affected in a negative way by large redundant data sets. This is the case, for instance, of \emph{Bayesian machine learning}. Bayesian machine learning provides a rigorous framework for performing inference over data. It allows the estimation of complete probability distributions, a precise evaluation of uncertainty, the possibility of neatly integrating prior expert knowledge in the learned model, and the ability to update the learned model when provided with new data. However, all these possibilities come at a high computational cost, as standard Bayesian learning relying on \emph{Markov chain Monte Carlo} (MCMC) algorithms do not to scale well with respect to the size of the data.

In presence of large redundant data sets, a possible solution to make inference via MCMC feasible consist in the reduction of the number of samples used to learn. Simple solutions include statistical techniques like \emph{random sampling} or unsupervised learning algorithms such as \emph{clustering via k-means} \cite{bishop2006pattern}. A more exact approach is based on the idea of creating \emph{coresets}: instead of learning on the whole redundant data sets, it may be possible to define a (weighted) subset of samples which is probabilistically guaranteed to return a result close to the one that would be obtained by processing the whole data. In Bayesian machine learning, \citet{campbell2017automated,campbell2018bayesian} recently proposed an efficient and promising algorithm to learn \emph{Bayesian coresets in Hilbert space} (BCH) that computes a weighted subset of samples by smartly exploiting the structure of that space.

As many other applications, network intrusion detection could take advantage of Bayesian data analysis. A careful estimation of uncertainty when evaluating the possibility of a threat on the network is critical in order to take decisions. The possibility of integrating expert knowledge and update models are also important features in the complex and constantly changing environment of networks. Unfortunately though, the amount of data collected by capturing packets on a network quickly exceeds the feasibility of performing Bayesian machine learning via MCMC. By filtering redundant data and reducing the amount of samples via BCH, network traffic data sets could be reduced in size (thus offering a concrete benefit for storing and management) and processed using Bayesian techniques (thus producing more complete and versatile results).

In this paper we conduct a preliminary study of the possibility of applying BCH to network intrusion detection data and perform Bayesian machine learning via MCMC. We consider two main problems. 
(i) \emph{How effective is the use of BCH to learn models of network intrusion?} We address this question considering (subsets of) realistic network traffic data and evaluating the effectiveness of BCH on a simple supervised learning problem. While an evaluation of BCH on computer security data (phishing data sets) is already provided in \citet{campbell2017automated,campbell2018bayesian} in terms of metrics of posterior quality, we offer here an analysis in terms of accuracy, which is more relevant to the field of cyber-security. In particular we consider our results in light of the trade-off between time-space and accuracy and with respect to the sensitivity of BCH to its hyper-parameters.
(ii) \emph{How effective is the use of BCH to reduce the amount of data in a streaming environment?} We answer this question by considering the same realistic network data, but setting up a more challenging scenario in which the data samples are received sequentially. In this context, we analyze, once again in terms of accuracy and time-space savings, the advantages that BCH may bring when processing the data in real-time, upon arrival. 
Our results confirms that the trade-off between accuracy and time-space savings when using BCH is mainly regulated by one of the free hyper-parameters of BCH. Moreover, we show that the algorithm could be successfully used in a streaming environment, where it succeeds in sensibly reducing the computational time over several iterations and in ensuring good performances by aggregating coresets over the same iterations. 

On the side, while tackling these questions, we also offer a practical contribution in the form of a porting of BCH algorithms\footnote{\url{https://github.com/trevorcampbell/bayesian-coresets}} \citep{campbell2017automated} into the framework of the probabilistic programming library Edward \citep{tran2016edward}. Specifically, we adapt the original code for coreset computation to work with Edward models, thus exploiting the probabilistic programming features of Edward\footnote{\url{http://edwardlib.org/}} and the automatic differentiation feature of Tensorflow\footnote{\url{https://www.tensorflow.org/}}. Code for this implementation is available online\footnote{\url{https://github.com/FMZennaro/BayesianCoresets-Edward}}.

The rest of the paper is organized as follows. Section \ref{sec:Background} briefly describes Bayesian machine learning and BCH. Section \ref{sec:NetworkSecurity} explains the problem of network intrusion detection. Section \ref{sec:ExperimentalSetup} introduces our experimental setup. Section \ref{sec:BCH-Base} tackles our first research question by analyzing the use of BCH on network traffic data. Section \ref{sec:BCH-Streaming} deals with our second research question by evaluating the use of BCH in a streaming environment. Finally, Section \ref{sec:Conclusion} summarizes our results and presents some of the several avenues available for further development of this work.

\section{Background \label{sec:Background}}

In this section we first introduce our general notation for the learning problem. We review the Bayesian approach to learning and its limitations. We then explain how Bayesian coresets deal with the problem of scalability. Finally, we review alternative approaches to work around the computational challenges of Bayesian learning. 

\subsection{Notation}
In the following, we will deal with standard \emph{supervised learning problems}. We consider a \emph{data matrix} $\dataset$ of dimension $\nsamples \times \nfeatures$, containing $\nsamples$ samples described by $\nfeatures$ features; a sample $\x$ is a vector of dimension $1 \times \nfeatures$. We also assume we are given a \emph{label vector}  $\labels$ of dimension $\nsamples$, such that for each sample $\x$ we have a label $\y$.
Our aim is to learn a model mapping samples to labels: $f_\param: \x \mapsto \y$, where $\param$ is a set of parameters defining the mapping function $f$.

The standard approach of machine learning is to convert this learning problem in an optimization problem as a function of the parameters $\param$. The optimal solution is found by computing the \emph{point} estimate $\hat{\paramI{}}$ of the parameters. For each input sample $\x$ we can then compute the output as $\y = f_{\hat{\param}} \left( \x \right)$. The result $\y$ (which can be interpreted probabilistically if calibrated \citep{shalizi2013advanced}) is the output of the single model $f_{\hat{\param}}$ on which we invested all our trust. 

\subsection{Bayesian machine learning}
In Bayesian machine learning we tackle the problem of supervised learning with the aim of computing a full \emph{distributional} estimation $P({\paramI{}})$ of the parameters $\param$, instead of a point estimation. In this way, for each input sample $\x$ we can compute a distribution over the possible outputs $P\left( \y \vert \x; \param \right)$. This result represents the probability distribution of the output, computed considering all possible values of the parameters $\param$ scaled by the trust assigned to them.

More formally, in Bayesian machine learning we estimate the \emph{posterior distribution} $\posterior{\param}=P\left(\param \vert \dataset\right)$ of the parameters given the data using \emph{Bayes' formula}:
\[
P\left(\param\vert \dataset\right)=\frac{P\left(\dataset \vert \param \right)P(\param)}{P(\dataset)} = \frac{P\left(\dataset \vert \param \right)P(\param)}{ \int P\left(\dataset \vert \param \right)P(\param) d\param } = \frac{ \likelihood{\dataset; \param} \prior{\param} }{ \int \likelihood{\dataset; \param} \prior{\param} d\param } ,
\]   
where $\prior{\param}=P(\param)$ is the \emph{prior probability distribution} over the parameters, $\likelihood{\dataset; \param}=P\left(\dataset \vert \param \right)$ is the \emph{likelihood function} of the data with respect to the parameters, and $P(\dataset)$ is the \emph{evidence}.

Computing the posterior distribution is a challenging task that requires the evaluation of the product of likelihood function $\likelihood{\dataset; \param}$ and prior distribution $\prior{\param}$, and the evaluation of the evidence integral. Monte Carlo Markov chain (MCMC) algorithms are a practical solution to this problem based on the idea of sampling from the posterior distribution \citep{givens2012computational}. The main drawback of this approach is the computational scalability as the complexity of sampling a posterior point grows linearly with the size of the data \citep{campbell2017automated}.

\subsection{Bayesian coresets \label{sec:BayesianCoresets}}
Evaluating the posterior distribution via MCMC sampling requires the computation of the likelihood $\likelihood{\dataset; \param}$. Under the assumption of independent and identically distributed data, the likelihood for the whole data set $\likelihood{\dataset; \param}$ may be factorized in the product of the likelihoods of individual data points $\likelihood{\x; \param}$:
\[
	\likelihood{\dataset; \param} = \prod_{i | x_i \in \dataset} \likelihood{\x; \param},
\]
or, equivalently, in the product of \emph{log-likelihoods}:
\[
	\log\likelihood{\dataset; \param} = \sum_{i | x_i \in \dataset} \log\likelihood{\x; \param}.
\]
Bayesian coresets compute a small weighted subset of the original data $\coreset \subseteq \dataset$ such that the log-likelihood computed on $\coreset$ approximates the log-likelihood  computed on $\dataset$:
\[
\lldata =
 \sum_{i | x_i \in \dataset} \log\likelihood{\x; \param} \approx \sum_{n | x_n \in \coreset} \weightI{n} \log\likelihood{\xI{n}; \param} = 
 \llcoreset,
\]
where $\xI{n}$ are samples belonging to the coreset and $\weightI{n}$ are the associated weights.
The degree of approximation may be evaluated in terms of $epsilon$-distance between the the original log-likelihood and the coreset likelihood:
\begin{equation} \label{eq:constrainedproblem}
	\left| \lldata - \llcoreset  \right| \leq
	\epsilon \cdot \left| \lldata \right| \,\,\, \forall \param.
\end{equation}
Estimating this distance is challenging, and an approximation is offered by \emph{Huggin's algorithm} \citep{huggins2016coresets}.

\paragraph{Bayesian coresets in Hilbert spaces.}
A refinement of this solutions has been proposed by \citet{campbell2017automated}, with the suggestion of embedding log-likelihoods in a Hilbert function space. This reformulation has several advantages. 
First, by taking the objects of this space to be functions of the form $g: \Theta \rightarrow \mathbb{R}$, log-likelihoods $\log \likelihood{\x; \param}$ or $\log \likelihood{\xI{n}; \weightI{n}, \param}$  become vectors of this space; consequently, the total likelihood over the whole data set $\lldata$ or the coreset $\llcoreset$ can be expressed in terms of vector sum.
Second, by taking as a norm of the space a bounded sup norm $\left\Vert g\right\Vert =\sup_{\param}\left|\frac{g\left(\theta\right)}{\lldata}\right|$, we can restate the constrained problem in Equation \ref{eq:constrainedproblem} as a sparse quadratic minimization problem:
\[ 
	\min_{\weights} \left\Vert 
	\log \likelihood{\dataset; \weights} - \log \likelihood{\dataset}
	 \right\Vert^2
\]
under the constraints:
\begin{align*}
& \weightI{n} \geq 0 &\\
& \sum \one{\weightI{n} > 0} \leq \maxcoresamples, & 
\end{align*}
where $\one{c}$ is the identity function that returns $1$ if $c$ holds or $0$ otherwise, and $M$ is a maximum number of coreset samples $\xI{n}$ that we allow selecting. This fomalization turns the problem of constructing a coreset into an optimization problem aimed at finding the minimal set of samples $\xI{n}$ that approximates the log-likelihood on the data set.
Finally, the structure of the Hilbert space allows us to exploit the directionality of the space in order to account for residual errors between $\log \likelihood{\dataset; \weights}$ and $\log \likelihood{\dataset}$ and to better select samples $\xI{n}$ that would improve the approximation.
Algorithms for coreset construction that exploit the properties of the Hilbert space include the \emph{coreset construction based on Frank-Wolfe algorithm} \citep{campbell2017automated} and the \emph{GIGA algorithm} \citep{campbell2018bayesian}.
  
\paragraph{Model dependency of Bayesian coresets.}
As it has been underlined by \citet{coleman2018select}, it is important to remark that a coreset computed by a BCH algorithm is tightly connected to a specific family of models. Such a coreset does not constitute a generic weighted non-redundant distillation of the original data set; it is a subset of the original data optimized with respect to a specific family of models in order to produce a posterior distribution as close as possible to the one that we would learn from the original data set. In sum, \emph{a BCH is actually a tuple made up by a family of models and a weighted set of samples}.   

\subsection{Alternative approaches to BCH for Bayesian learning}
BCH is just one of the possible approaches to make Bayesian machine learning feasible on large data sets. Other approaches which do not involve reducing the number of samples include \emph{variational Bayes}, \emph{parallel MCMC} and \emph{approximate MCMC}. 
Variational Bayes algorithms forgo the idea of using MCMC algorithms to perform inference, and rely instead on variational approximations of the posterior \citep{bishop2006pattern}. The variational approach allows learning in presence of large data sets, but the method does not provide guarantees on the degree of approximation of the uncertainty of the posterior \citep{giordano2015linear}.
Parallel MCMC algorithms rely on parallelization: large data sets are divided among multiple clusters; each cluster runs locally Bayesian inference via MCMC and produces a posterior distribution; finally, all the posteriors are aggregated by finding a unique posterior in the metric space of the posterior distributions \citep{neiswanger2013asymptotically, srivastava2015wasp}. The parallel approach allows to deal with large data sets, but it still requires a high computational budget and does not address the problem of storing redundant data.
Finally, approximate MCMC aims at speeding up existing algorithms by replacing costly transition in the Markov chain process with approximations \citep{johndrow2015approximations}. Again, this approach is effective when we have to process large data sets, but it requires analyzing the execution of the MC algorithm, and, once again, it does not consider the problem of storing redundant data.

\section{Network Security \label{sec:NetworkSecurity}}
Network security is one of the main challenges in the management of online systems. Network administrators try to monitor and prevent malicious activity through the deployment of intrusion detection systems, the collection of network traffic, and the analysis of this data \citep{northcutt2002network}. Processing these data in a timely manner in order to detect suspicious activity as early as possible is a crucial problem. The ease with which large amount of data can be collected on a network poses severe scalability problems, both in terms of storage and in terms of processing \citep{gardiner2016security}. Given this constraint, computationally-cheap algorithms, such as signature-matching, random forests and support vector machines, have been favored; for a review of pattern recognition and classical machine learning algorithms applied to network security, see, for instance, \citet{gardiner2016security} and \citet{garcia2009anomaly}.

\section{Experimental Setup \label{sec:ExperimentalSetup}}

In this section we provide a formal description of our study by defining the exact learning problem we considered, by presenting the data sets and the transformations we applied to them, and, finally, by discussing the models we implemented.

\paragraph{Problem definition}
Given a large data set for network intrusion detection, we express our learning problem as a supervised learning problem in which we try to discover a function that maps network flows to an output defining whether a flow is malicious or not. More precisely, we try to infer an optimal set of parameters $\theta$ that define the mapping function $f_\param: \x \mapsto \y$.
In a first static scenario, we process our data with and without BCH, running the simulations multiple times ad observing the contribution of the BCH algorithm.
In a second scenario, we simulate the progressive collection of large chunks of data. All the data samples are taken to be independent and identically distributed. In this case, we observe what the contribution of BCH would be if we were to filter out data as soon as they are collected. 

\paragraph{Network data set.}
To run our experiments we use the network traffic data collected in the CICIDS2017 data set \citep{sharafaldin2018toward}. 
Processing real-world network data presents challenges from a privacy perspective; for this reason, the CICIDS2017 was collected running a simulated network designed to behave in a realistic fashion. Five days of simulated traffic were collected; during each day, different types of attacks and malicious behaviors were enacted. Network packets are gathered and aggregated in network flows. In total, the data set contains more than 2.5 million samples. Each sample is defined by a $78$-dimensional vector reporting features such as packet flags and packet lengths. Finally, a binary label has been assigned to each network flow, denoting whether a flow is legitimate or not. We restrict our attention to the second day, Tuesday\footnote{Notice that we did not consider the data on Monday because no malicious activity takes place on this day, thus providing us only with positive instances.}, which is made up by 445708 data samples, of which 13835 constitute instances of brute force attacks. 

\paragraph{Network data preprocessing.}
The CICIDS2017 data set contains a very limited number of samples (201) with missing values for the day of Tuesday. Given their limited number we simply assume that they are \emph{missing completely at random} \citep{barber2012bayesian} and we just drop them.
Also, before each experiment we always standardize the data to zero mean and unit variance by feature. Standardization parameters are always computed exclusively on the training data being processed. 

\paragraph{Data set subsampling.}
For the purpose of this preliminary study, we evaluate our algorithms only on limited subsets of the large CICIDS2017 data set. Studying smaller data sets guarantees some advantages: (i) it allows us to compare Bayesian learning on coresets with the "ground truth" of learning on the whole data sets, which would not be feasible if we were to consider the entire data set; (ii) manipulating the data set allows us to simulate scenarios in which we receive streaming data. While we plan to extend our evaluation to the full data set in order to assess the true potential of BCH for network intrusion detection, we were still able to get useful insights by studying its application on more modest-sized subsets of the original data set.
Thus, from the large original pool of data, we programmatically create subsets by random sampling. In order to preserve the imbalance between positive and negative instances in the training data, we always select ten times as many positive samples as negative samples. The prototypical training data set $\traindata{i}$ consists of 800 positive instances and 80 negative instances. For a test data set, instead, we selected an even number of positive and negative instances, thus simplifying the interpretation of the results. Our prototypical test data set $\testdata{i}$ consists of 200 positive an 200 negative samples.

\paragraph{Sample reduction techniques.}
Given a training data set $\traindata{i}$, in order to reduce the amount of data samples, we apply BCH using the GIGA algorithm \citep{campbell2018bayesian}. This algorithm has two free hyper-parameters: (i) the number of random dimension on which to project the samples; and (ii) the number of computational iteration $M$, which implicitly limits the maximum number of coreset samples that can be selected.

\paragraph{Network models.}
For our simulations, we consider two models:
\begin{enumerate}
	\item \emph{Bayesian logistic regression (BLR):} a discriminative generalized-linear Bayesian machine learning algorithm \citep{bishop2006pattern}. 
	We define our weighted BLR model as:
	\begin{align*}
	\theta & = \gaussianpdf{0}{1} \\
	\y & = \bernoullipdf{ \left( \logistic{\param \x} \right)^{\weightI{i}} }
	\end{align*}
	where $\gaussianpdf{0}{1}$ is a Gaussian prior, and the likelihood $\likelihood{\dataset;\param}$ is the likelihood under a Bernoulli pdf $\bernoullipdf{ \logistic{\param \x}}$ scaled by the weight $\weightI{i}$ associated to the sample $\x$. If the data samples are not weighted, then $\weightI{i}=1$ for all $\x$, and the model reduces to a standard BLR. Notice that, in general, if all the samples $\x$ are scaled by a constant value $\weightI{i}=k$ the inference process will not change\footnote{We take advantage of this constant scaling to prevent overflowing errors in the simulations.}.
	
	Given a new sample $\x$, its probability of belonging to a class $\y$ can be obtained by integrating over all the models under the posterior:
	$
	P(\y | \x) = \int P(\y | \x, \param) P(\param) d\param.
	$	 
		
	\item \emph{Support Vector Machine (SVM):} as a baseline and comparison, we consider support vector machine, a linear maximum-margin discriminator \citep{cortes1995support}. We train a SVM model to find the slope $\param$ of a discriminating hyper-plane between samples\footnote{Notice that for a fair comparison with the BLR model, we compute only the slope of the discriminating hyperplane and not its intercept.}.
	
	Given a new samples $\x$, its class is computed as $\y = \sign{\param \x}$, where $\sign{z}$ is the sign function, $\sign{z} = 1$ if $z>0$, $0$ otherwise.
\end{enumerate}

\section{Simulation 1: BCH Applied to Network Intrusion Detection Data \label{sec:BCH-Base}}

In this simulation we analyze the use of BCH applied to network intrusion detection data. We evaluate the contribution they provide both from the point of view of the performance they achieve and the time-space they save. We compare these results to the baseline offered by SVM and by a BLR computed over the whole data set.

\paragraph{Protocol.}
We generate five subsets of training data $\traindata{i}$ and test data $\testdata{i}$, $1 \leq i \leq 5$, using the methodology presented in Section \ref{sec:ExperimentalSetup}.
 
We apply BCH to each training data set $\traindata{i}$. We set the free hyper-parameters of BCH as follows: (i) we fix the number of random dimension to $500$, following the experimental evaluation in \citet{campbell2017automated}; (ii) we consider three values for the number of computational iteration $M$, that is $M=1000$, following again the evaluation in \citet{campbell2017automated}, $M=500$, and an aggressively lower value of $M=100$, which is expected to guarantee a higher saving in terms of space and time.

For each subset $i$, we train and test an SVM model, a BLR model trained on the whole data set $\traindata{i}$, and a BLR model trained on the coreset computed from the training data $\traindata{i}$.\footnote{Notice that we do not train the SVM model on the coresets because, as discussed in Section \ref{sec:BayesianCoresets}, coresets are not generic non-redundant sub-data sets, but they are sub-selections optimized for a specific statistical model.} The SVM model is trained with default parameters from the scikit\footnote{\url{https://scikit-learn.org/stable/}} library. The BLR models are trained using the Hamiltonian Monte Carlo algorithm offered in the Edward library with the following settings: sampling 10000 points, using a burn-in period of half of the samples, thinning every second sample, and adjusting the step size manually to guarantee an acceptance rate around 0.8. When doing prediction, we use 1000 posterior samples.
We repeat each training and testing 10 times and we average the results.

We evaluate the results in terms of classification accuracy and wall-clock time required for the training of the model (all the models are run on a non-dedicated mid-range laptop machine with no GPU support).

\paragraph{Results.}
First of all, the data processing via BCH with different hyper-parameters produced different coresets. Table \ref{tab:coresets} reports the number of data points selected, specifying the number of samples in the minority class that have been preserved, and the wall-clock computation time as a function of the hyper-parameter $M$.
Notice that the number of iterations $M$ does not correspond to the number of coreset samples selected; in all the cases the algorithm selects a number of samples well below this threshold. With respect to the original number of samples, the amount of data points selected by BCH ranges from around one tenth, when using a low $M=100$, to one third, when using a high $M=1000$.  

\setlength{\tabcolsep}{5pt}
\begin{table}
	\begin{centering}
		\begin{tabular}{ccccccc}
			\hline 
			\textbf{Hyper-parameter} & $\traindata{1}$ & $\traindata{2}$ & $\traindata{3}$ & $\traindata{4}$ & $\traindata{5}$ & \textbf{Time(s)} \tabularnewline 
			\hline 
			$M=100$ & {\footnotesize{}84/1} & {\footnotesize{}82/3} & {\footnotesize{}82/1} & {\footnotesize{}88/1} & {\footnotesize{}87/3} & {\footnotesize{}$260.78\pm2.37$}\tabularnewline
			\hline 
			$M=500$ & {\footnotesize{}184/4} & {\footnotesize{}191/10} & {\footnotesize{}186/7} & {\footnotesize{}189/9} & {\footnotesize{}200/10} & {\footnotesize{}$251.82\pm3.05$}\tabularnewline
			\hline 
			$M=1000$ & {\footnotesize{}259/5} & {\footnotesize{}270/15} & {\footnotesize{}239/4} & {\footnotesize{}252/8} & {\footnotesize{}252/6} & {\footnotesize{}$268.63\pm2.51$}\tabularnewline
			\hline 
		\end{tabular}
		\par\end{centering}
	\caption{Coresets computed on each training data set $\traindata{i}$. The table reports the number of data points selected to the left of the slash (/), and the number of these points belonging to the minority class to the right of the slash (/). The last column reports average and standard deviation of the wallclok time to compute the coresets. \label{tab:coresets}}
\end{table}

\begin{figure}
	\centering
	\includegraphics[scale=0.5]{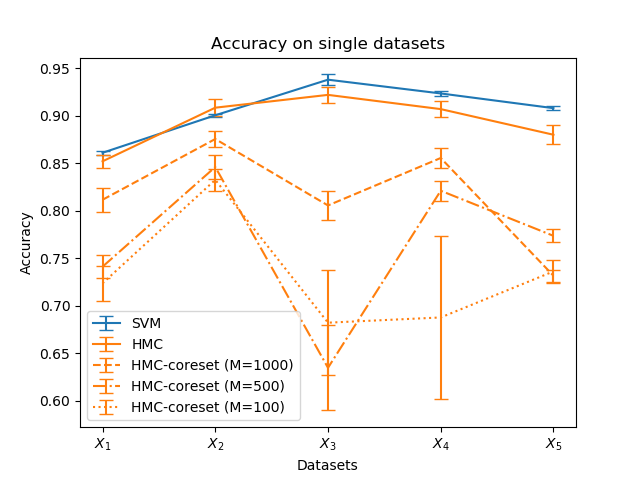}
	\caption{Mean and standard deviation of accuracy of the models on each data set $\testdata{i}$. \label{fig:accuracy1}}
\end{figure}

Figure \ref{fig:accuracy1} shows the accuracy of our models on the different data sets we considered. Consistently with our expectations, the two linear models, SVM and BLR on the whole data set, perform similarly; BLR models trained on coresets show, in general, decreasing performances as we decreased the hyper-parameter $M$ of the BCH algorithm.

\begin{figure}
	\centering
	\includegraphics[scale=0.5]{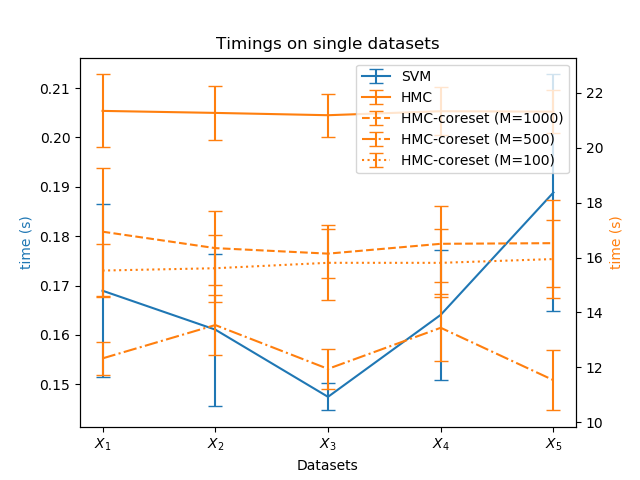}
	\caption{Mean and standard deviation of wall-clock time required for training the models on each data set $\traindata{i}$. Notice the different y-scale for SVM on the left side, and the BLR model on the right side. \label{fig:time1}}
\end{figure}

Figure \ref{fig:time1} compares the wall-clock time of each algorithm. The highly optimized SVM algorithm shows some variability, but in general terminates in tenths of seconds. On the other hand, BLR takes up to two orders of magnitude longer. BLR on coresets is faster, even if on the same time scale; surprisingly using coresets with $M=500$ took the shortest time, which may be due to the particularly good sub-selection of points, or, more likely, to other contingent processes running on the same machine. 

\paragraph{Discussion.}
These basic experiments highlight that the time-space savings offered by BCH inevitably come at the cost of the accuracy of the final model. The number of iterations $M$ provides a key hyper-parameter to manage such a trade-off, as it exchanges the dimension of the data set for the accuracy of the model.

Notice that from these experiments, the time saving offered by coresets does not appear particularly remarkable. It is worth, though, to underline that such an improvement is relevant when related to the small data sets we are processing. The time savings when using larger data set are discussed in detail in \citet{campbell2017automated}.

Interestingly, the computation of coresets has a subsampling effect with respect to the minority class, as shown in Table \ref{tab:coresets}: while in the original data set the ratio between the two classes was set to 1:10, this ratio has sensibly decreased. This may seem undesirable if we were expecting BCH to produce a more balanced data set; in reality, though, the algorithm selects only samples useful for a proper reconstruction of the likelihood function, and the result seems to suggest that the instances of malicious behaviours may actually be quite redundant, probably due to the fact that we are considering only one specific form of attack (brute force).

\section{Simulation 2: BCH in a Streaming Environment \label{sec:BCH-Streaming}}

In this simulation we try to setup a more interesting and realistic scenario. We simulate the collection of batches of data in real-time and we learn from the cumulative set of collected samples. The aim is to evaluate how the learning process would be affected if the sets of collected data were to be downsized using BCH before being processed and stored. Such a scenario seems particularly interesting because BCH would immediately discard redundant data, thus solving at once the problem of making Bayesian inference feasible and reducing the required amount of memory and storage space.

\paragraph{Protocol.}
As before, we generate five subsets of training data $\traindata{i}$ and test data $\testdata{i}$, $1 \leq i \leq 5$. Now, however, instead of processing each data set independently, we simulate the arrival of a data set $\traindata{i}$ at time steps $t_i$. At each time step $t_i$, we want to learn from all the collected data sets and so we pool together all the data $\traindata{j}$, for $j \leq i$. Notice that all the samples are independent and identically distributed.

When using coresets, we apply BCH to each training data set $\traindata{i}$ as soon as it is collected. At each time step $t_i$, instead of aggregating together all the previously collected data, we just aggregate the coresets. This operation is theoretically justified by the possibility of aggregating coreset computed in parallel \citep{campbell2017automated}. We use the same hyper-parameters for BCH used in the previous simulation.  

We also run the same models as before, and we repeat each simulation 10 times.

\paragraph{Results.}
We work with the same coresets computed in the previous simulation and we refer back the reader to Table \ref{tab:coresets} for their details.

\begin{figure}
	\centering
	\includegraphics[scale=0.5]{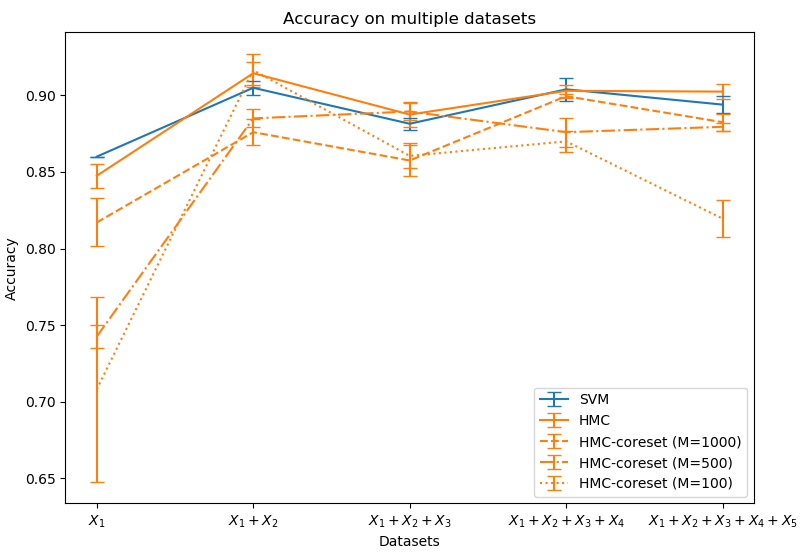}
	\caption{Mean and standard deviation of accuracy of the models on each data set $\testdata{i}$. \label{fig:accuracy2}}
\end{figure}

Figure \ref{fig:accuracy2} shows the accuracy of the models on the different data sets we generated. The starting values of accuracy computed on a single data set ($X_1$) are consistent with the values computed in the previous experiments and shown in Figure \ref{fig:accuracy1}. When we start aggregating more data sets, we notice that the performance of SVM and HMC on the whole data set is only slightly improved; on the other hand, the performance of BLR on coresets shows a consistent improvement. Even aggregating only two coresets ($X_1+X_2$) the performance gap between BLR on coresets and SVM or BLR on the whole data set is significantly reduced. This improvement is expected when aggregating two coresets computed with the hyper-parameter $M=500$; in this case, the final amount of selected data points would be close to the amount obtained computing a single coreset with the hyper-parameter $M=1000$; and we know from the previous simulations that the performance of BLR trained on a coreset computed with the hyper-parameter $M=1000$ is very close to the performance of BLR on the whole data set. More surprising is the improvement registered by aggregating only two coresets computed with hyper-parameter $M=100$. 
 
Figure \ref{fig:time1} compares the wall-clock time of each algorithm. Again, the time scale of the two family of algorithms, SVM and BLR, are very different. However, notice that while the computational time for SVM and BLR on the whole data sets tend to grow in a linear fashion, the growth in the required computational time when running BLR on coresets is almost flat.
\begin{figure}
	\centering
	\includegraphics[scale=0.5]{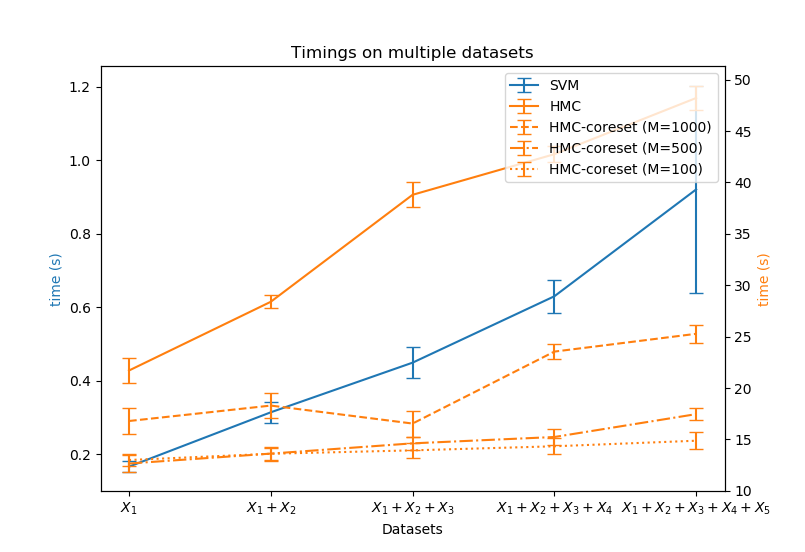}
	\caption{Wall-clock time required for training the models on each data set $\traindata{i}$. Notice the different y-scale for SVM on the left side, and the BLR model on the right side. \label{fig:time2}}
\end{figure}

\paragraph{Discussion.}
This simulation showed the potential advantages that could be obtained by deploying BCH in a streaming scenario. In such an instance, the aggregation of two or more coresets can provide a performance very close to SVM or BLR trained on the whole data. One of the most significant advantages, though, is that BCH reduces the amount of data in real-time before learning, thus limiting the amount of memory necessary for processing; guaranteeing a sub-linear growth in the time required for learning as more data are gathered may prove especially advantageous when data is collected in real-world environments in which batches of data are generated continuously over multiple timesteps.

\section{Conclusion and Future Work \label{sec:Conclusion}}
This preliminary study showed the feasibility of applying BCH to network data. Network intrusion detection could take great advantage by employing fully probabilistic descriptions of network traffic, and BCH may prove to be an enabler for such an approach. Moreover, we showed that the same algorithm is also effective in reducing the number of samples to be stored; this issue is particularly relevant when collecting network packets, as the amount of data may quickly grow and cause severe challenges in their management. 

Our experiments first confirmed the concrete trade-off between model accuracy and time-space saving when adopting BCH. More interestingly, we also investigated how BCH may be deployed in a dynamic streaming environment in which data samples would be filtered in real-time before processing. This last scenario returned particularly good and interesting results, showing that BCH can be effectively used to sub-select relevant data samples at different time steps and aggregate together only the coresets. We demonstrated that in a streaming scenario the use of BCH may guarantee a better scalability by ensuring that the computational time for learning grows in a strongly sub-linear fashion. 

Of course this study is just a preliminary evaluation of the potential of BCH applied to the challenging problem of processing the large data sets for network security. Further investigation is clearly necessary to assess more precisely the role that BCH may serve. Immediate directions of further study that we consider are the following: applying our protocol to bigger and more realistic data sets; include other typologies of attacks; compare BCH to other data reduction techniques, such as random sampling or k-means. More interesting questions concern also the recursive application of BCH in a streaming scenario and its effectiveness when used to process streaming data that do not conform to the assumption of independent and identically distributed data anymore. 

\bibliographystyle{plainnat}
\bibliography{../../lib/lib}

\begin{thebibliography}{19}
\providecommand{\natexlab}[1]{#1}
\providecommand{\url}[1]{\texttt{#1}}
\expandafter\ifx\csname urlstyle\endcsname\relax
  \providecommand{\doi}[1]{doi: #1}\else
  \providecommand{\doi}{doi: \begingroup \urlstyle{rm}\Url}\fi

\bibitem[Barber(2012)]{barber2012bayesian}
David Barber.
\newblock \emph{Bayesian reasoning and machine learning}.
\newblock Cambridge University Press, 2012.

\bibitem[Bishop(2006)]{bishop2006pattern}
Christopher~M Bishop.
\newblock \emph{Pattern recognition and machine learning}.
\newblock springer, 2006.

\bibitem[Campbell and Broderick(2017)]{campbell2017automated}
Trevor Campbell and Tamara Broderick.
\newblock Automated scalable bayesian inference via hilbert coresets.
\newblock \emph{arXiv preprint arXiv:1710.05053}, 2017.

\bibitem[Campbell and Broderick(2018)]{campbell2018bayesian}
Trevor Campbell and Tamara Broderick.
\newblock Bayesian coreset construction via greedy iterative geodesic ascent.
\newblock \emph{arXiv preprint arXiv:1802.01737}, 2018.

\bibitem[Coleman et~al.(2018)Coleman, Mussmann, Mirzasoleiman, Bailis, Liang,
  Leskovec, and Zaharia]{coleman2018select}
Cody Coleman, Stephen Mussmann, Baharan Mirzasoleiman, Peter Bailis, Percy
  Liang, Jure Leskovec, and Matei Zaharia.
\newblock Select via proxy: Efficient data selection for training deep
  networks.
\newblock 2018.

\bibitem[Cortes and Vapnik(1995)]{cortes1995support}
Corinna Cortes and Vladimir Vapnik.
\newblock Support-vector networks.
\newblock \emph{Machine learning}, 20\penalty0 (3):\penalty0 273--297, 1995.

\bibitem[Garcia-Teodoro et~al.(2009)Garcia-Teodoro, Diaz-Verdejo,
  Maci{\'a}-Fern{\'a}ndez, and V{\'a}zquez]{garcia2009anomaly}
Pedro Garcia-Teodoro, Jesus Diaz-Verdejo, Gabriel Maci{\'a}-Fern{\'a}ndez, and
  Enrique V{\'a}zquez.
\newblock Anomaly-based network intrusion detection: Techniques, systems and
  challenges.
\newblock \emph{computers \& security}, 28\penalty0 (1-2):\penalty0 18--28,
  2009.

\bibitem[Gardiner and Nagaraja(2016)]{gardiner2016security}
Joseph Gardiner and Shishir Nagaraja.
\newblock On the security of machine learning in malware c\&c detection: A
  survey.
\newblock \emph{ACM Computing Surveys (CSUR)}, 49\penalty0 (3):\penalty0 59,
  2016.

\bibitem[Giordano et~al.(2015)Giordano, Broderick, and
  Jordan]{giordano2015linear}
Ryan~J Giordano, Tamara Broderick, and Michael~I Jordan.
\newblock Linear response methods for accurate covariance estimates from mean
  field variational bayes.
\newblock In \emph{Advances in Neural Information Processing Systems}, pages
  1441--1449, 2015.

\bibitem[Givens and Hoeting(2012)]{givens2012computational}
Geof~H Givens and Jennifer~A Hoeting.
\newblock \emph{Computational statistics}, volume 710.
\newblock John Wiley \& Sons, 2012.

\bibitem[Huggins et~al.(2016)Huggins, Campbell, and
  Broderick]{huggins2016coresets}
Jonathan Huggins, Trevor Campbell, and Tamara Broderick.
\newblock Coresets for scalable bayesian logistic regression.
\newblock In \emph{Advances in Neural Information Processing Systems}, pages
  4080--4088, 2016.

\bibitem[Johndrow et~al.(2015)Johndrow, Mattingly, Mukherjee, and
  Dunson]{johndrow2015approximations}
James~E Johndrow, Jonathan~C Mattingly, Sayan Mukherjee, and David Dunson.
\newblock Approximations of markov chains and high-dimensional bayesian
  inference.
\newblock \emph{arXiv preprint arXiv:1508.03387}, 2015.

\bibitem[LeCun et~al.(2015)LeCun, Bengio, and Hinton]{lecun2015deep}
Yann LeCun, Yoshua Bengio, and Geoffrey Hinton.
\newblock Deep learning.
\newblock \emph{nature}, 521\penalty0 (7553):\penalty0 436, 2015.

\bibitem[Neiswanger et~al.(2013)Neiswanger, Wang, and
  Xing]{neiswanger2013asymptotically}
Willie Neiswanger, Chong Wang, and Eric Xing.
\newblock Asymptotically exact, embarrassingly parallel mcmc.
\newblock \emph{arXiv preprint arXiv:1311.4780}, 2013.

\bibitem[Northcutt and Novak(2002)]{northcutt2002network}
Stephen Northcutt and Judy Novak.
\newblock \emph{Network intrusion detection}.
\newblock Sams Publishing, 2002.

\bibitem[Shalizi(2013)]{shalizi2013advanced}
Cosma Shalizi.
\newblock Advanced data analysis from an elementary point of view, 2013.

\bibitem[Sharafaldin et~al.(2018)Sharafaldin, Lashkari, and
  Ghorbani]{sharafaldin2018toward}
Iman Sharafaldin, Arash~Habibi Lashkari, and Ali~A Ghorbani.
\newblock Toward generating a new intrusion detection dataset and intrusion
  traffic characterization.
\newblock In \emph{ICISSP}, pages 108--116, 2018.

\bibitem[Srivastava et~al.(2015)Srivastava, Cevher, Dinh, and
  Dunson]{srivastava2015wasp}
Sanvesh Srivastava, Volkan Cevher, Quoc Dinh, and David Dunson.
\newblock Wasp: Scalable bayes via barycenters of subset posteriors.
\newblock In \emph{Artificial Intelligence and Statistics}, pages 912--920,
  2015.

\bibitem[Tran et~al.(2016)Tran, Kucukelbir, Dieng, Rudolph, Liang, and
  Blei]{tran2016edward}
Dustin Tran, Alp Kucukelbir, Adji~B. Dieng, Maja Rudolph, Dawen Liang, and
  David~M. Blei.
\newblock {Edward: A library for probabilistic modeling, inference, and
  criticism}.
\newblock \emph{arXiv preprint arXiv:1610.09787}, 2016.

\end{thebibliography}

\end{document}